\documentclass{article}

\usepackage{PRIMEarxiv}
\usepackage{comment}
\usepackage[utf8]{inputenc} 
\usepackage[T1]{fontenc}    
\usepackage{hyperref}       
\usepackage{url}            
\usepackage{booktabs}       
\usepackage{amsfonts}       
\usepackage{nicefrac}       
\usepackage{microtype}      
\usepackage{lipsum}
\usepackage{float}
\usepackage{subcaption}
\usepackage{fancyhdr}       
\usepackage{graphicx}       
\graphicspath{{media/}}     

\title{Extended OpenTT games dataset: A table tennis dataset for fine-grained shot type and point outcome detection}   
\author{
  Moamal Fadhil Abdul\textendash Mahdi \\
  DTU Compute \\
  2800 Kgs. Lyngby, Denmark \\
  \texttt{moamal001@gmail.com}
  \And
  Jonas Bruun Hubrechts \\
  DTU Compute \\
  2800 Kgs. Lyngby, Denmark \\
  \texttt{jbrhu@dtu.dk}
  \And
  Thomas Martini Jørgensen\thanks{Shared last authorship.} \\
  DTU Compute \\
  2800 Kgs. Lyngby, Denmark \\
  \texttt{tmjq@dtu.dk}
  \And
  Emil Hovad\footnotemark[1] \\
  DTU Compute \\
  2800 Kgs. Lyngby, Denmark \\
  \texttt{emilh@dtu.dk}
}

\begin{document}
\maketitle

\begin{abstract}
Automatically detecting and \emph{classifying} strokes in table tennis video can streamline training workflows, enrich broadcast overlays, and enable fine-grained performance analytics. For this to be possible, annotated video data of table tennis is needed. We extend the public OpenTTGames dataset with highly detailed, frame-accurate \textbf{shot type} annotations (forehand, backhand with subtypes), \textbf{player posture} labels (body lean and leg stance), and \textbf{rally outcome} tags at point end. OpenTTGames is a set of recordings from the side of the table with official labels for bounces, when the ball is above the net, or hitting the net. The dataset already contains ball coordinates near events, which are either "bounce", "net", or "empty\_event" in the original OpenTTGames dataset, and semantic masks (humans, table, scoreboard). Our extension adds the types of stroke to the events and a per-player taxonomy so models can move beyond event spotting toward tactical understanding (e.g., whether a stroke is likely to win the point or set up an advantage). We provide a compact coding scheme and code-assisted labeling procedure to support reproducible annotations and baselines for fine-grained stroke understanding in racket sports. This fills a practical gap in the community, where many prior video resources are either not publicly released or carry restrictive/unclear licenses that hinder reuse and benchmarking. Our annotations are released under the same \href{https://creativecommons.org/licenses/by-nc-sa/4.0/}{CC BY-NC-SA 4.0} license as OpenTTGames, allowing free non-commercial use, modification, and redistribution, with appropriate attribution.
\end{abstract}


\section{Introduction}\label{sec:introduction}
In competitive racket sports, video analysis is a crucial tool for refining technique and developing strategy. Coaches and players meticulously break down match footage to identify tactical patterns and biomechanical flaws. However, this manual process is often time-consuming and subjective, creating a bottleneck for high-volume analysis. To address this, machine learning techniques are increasingly being used to automate and enhance the analytical process. 

A primary objective in this domain is the automated detection and classification of player strokes. Such a capability is fundamental for modern performance analysis, allowing coaches and players to query specific aspects of game play. For instance, they could instantly review all forehand shots down the line or analyze rally dynamics, thereby accelerating skill improvement and strategic development.

Beyond automated performance analysis, one can also transform the real-time viewing experience for audiences. By automatically identifying strokes during live broadcasts, systems can provide enriched visualizations, insightful data-driven commentary, and interactive statistics. This makes the sport more engaging and accessible to both casual viewers and seasoned fans alike.

A variety of machine learning approaches have been applied to table tennis video analysis. Foundational work such as TTNet\cite{Voeikov_2020_CVPR_Workshops} focused on fundamental event spotting, using a multi-task encoder–decoder architecture to detect ball bounces and net hits from a sideline view, providing segmentation masks as a key output. Other research has shifted the focus to the players themselves. RallyTempPose\cite{A_Stroke_of_Genius}, for example, pioneered the prediction of future strokes by analyzing player skeletons, using a dual attention mechanism to capture game dynamics and opponent interactions. More directly related to stroke classification, researchers have explored various deep learning architectures. In the analogous domain of tennis, for example, a model such as SlowFast \cite{feichtenhofer2019slowfastnetworksvideorecognition} exemplifies the effectiveness of temporal deep learning in general and is applied in \cite{hovad2024classificationtennisactions} on the THETIS benchmark dataset \cite{gourgari2013thetis}. 

For table tennis, ViSTec\cite{vistec} represents a notable work, achieving high accuracy by leveraging a powerful VideoMae\cite{VideoMAE} transformer model combined with a graph for contextual understanding. With regard to fine-grained stroke classification, this has been heavily driven by the introduction of specialized datasets. The most notable of these is TTStroke-21 \cite{martin:hal-02937666}, introduced as part of the MedieEval benchmarking initiative. With its annotations for 20 distinct stroke types, it now serves as the standard dataset for the associated benchmark test. In a different approach focusing on capturing high-fidelity motion, Kulkarni and Shenoy collected a custom dataset \cite{kulkarni2021tabletennis}. Rather than using typical match footage, they recorded an isolated player from a head-on camera angle, arguing this provides a clearer view of stroke biomechanics for the 11 types they annotated. 
Some recent work has explored classifying strokes using only the ball's trajectory data — a novel, minimalist approach that sidesteps visual analysis of the player entirely \cite{kulkarni2023tabletennisstrokedetection}. While these works demonstrate significant value, a persistent challenge is data accessibility. The specialized datasets used in many of these studies, such as the one introduced by Kulkarni and Shenoy and the trajectory data for motion-based analysis, are not publicly released. Moreover, while benchmark datasets like TTStroke-21 are available for specific tasks, they often carry restrictive licenses that hinder broader reuse. This lack of a large-scale, openly licensed dataset with rich tactical labels limits reproducible research and comparative benchmarking.

To fill this gap, this study presents an extensive enhancement of the annotations in the public dataset known as OpenTTGames. We manually annotated all matches in the dataset, yielding detailed pose at stroke instances. Our annotations provide a multi-layered view of the game, moving beyond simple stroke labels to include crucial tactical context. For each stroke, we record the player executing it with their table position (left/right), the stroke type with its subcategories, and two biomechanical markers: body lean and leg stance. Additionally, we create a separate set of annotations for rally-ending events, by tagging the conclusion of each rally with outcome labels — such as winner or out — and attributing the result to a specific player. While these annotations are temporally aligned with the final stroke of a rally and precede the first stroke of the subsequent rally, they are treated as a distinct layer and can be used independently of the stroke annotations. This process also refines the existing event data in OpenTTGames; for example, some net hits are now explicitly attributed to the player who made the error, and certain bounce events are now identified as double bounces. By adding this rich, frame-accurate layer of tactical and biomechanical data to a widely accessible public resource, we aim to enable the development of models that can move beyond event spotting toward a deeper, tactical understanding of the sport and inspire more open-source datasets in the racket sport domain.

\section{Data}
\label{sec:data}
In this section, we describe the additions we made to the original OpenTTGames dataset\cite{Voeikov_2020_CVPR_Workshops}.
The code for accessing the extended data is provided in the link \url{https://github.com/moamal01/table_tennis_data}.

\subsection{Original OpenTTGames}
\label{sec:opentt}
OpenTTGames provides a rich set of existing annotations, including ball bounces and ball positions at key moments, and our work extends this public dataset. The original dataset consists of 12 video files split into five training videos and seven test videos. Importantly, the training videos are not necessarily full-length matches, and some of them include warm-up segments. The test videos are significantly shorter. Only three left-handed players are featured, all of them in the test videos. All videos were recorded from a single, fixed side of the table at 1920 × 1080 resolution and 120 frames per second (fps) and the videos do not contain an audio track. The static camera perspective ensures that both players are consistently visible (when not too far from the table), while the high frame rate is crucial for analyzing the sport's rapid movements. Crucially for our work, the original dataset also marks the timestamp of every racket-ball impact. However, these events are labeled only as untyped "empty\_event", marking the instant of racket–ball contact without any stroke-type classification. They appear in the first three training videos as well as the fifth. Despite only being simple timestamps as opposed to full annotations, they reduced the annotation time and provided guidance for timestamping the other videos.

\subsection{Annotations}
In order to make our dataset applicable to a wide variety of use cases, we decided to make the stroke annotations as fine-grained as possible by: (i) specifying the side of the table, (ii) distinguishing between forehand and backhand strokes, and (iii) identifying the exact stroke technique used.
Due to the importance of the player's balance during a stroke, we also incorporate body lean information (forward, backward, right, and left). Similarly, we also annotate which foot is lifted in the moment of a stroke. The annotations are summarized in the overall table of the data \ref{tab:overall}.

\begin{table}[H]
\centering
\caption{Data summary and enrichment overview.}
\begin{tabular}{llll}
\toprule
\textbf{ } & \textbf{Training} & \textbf{Test} & \textbf{Total} \\ 
\midrule
Videos  & 5 & 7 & 12 \\
Right handed players & 12 & 10 & 22\\ 
Left handed players & 0 & 4 & 4 \\
\bottomrule
Strokes annotated & 1,134 & 323 & 1,457 \\ 
Lean information annotated & 1,117 & 315 & 1,432 \\ 
Feet information annotated & 1,025 & 294 & 1,319 \\
\bottomrule
Rally endings annotated & 223 & 58 & 282 \\ 
\bottomrule
\end{tabular}
\label{tab:overall}
\end{table}

It is important to note that the training/test split originated from the OpenTTGames dataset and is not strictly meaningful for our dataset. Because some fine-grained stroke types do not appear in the test videos and vice versa, the original split does not align with the extended dataset’s intended use cases. Users are therefore encouraged to apply their own splits. For simplicity and to maintain a clear overview of our additions, we have nonetheless kept the original split.

\section{Method}
We enriched the data in OpenTTGames using a custom annotation script that allowed frame-by-frame inspection and labeling. Labels were entered using a compact shorthand notation that expanded into predefined label parts, ensuring consistent naming and reducing annotation time and typos. The script also allowed the insertion of temporary tags (e.g., point) for later review and refinement. An overview of the shorthand notation is provided in \autoref{sec:codes}.

The overall procedure for adding stroke and rally-ending timestamps followed a two-step process.
\begin{itemize}
    \item Step 1 consisted of identifying and tagging all stroke frames with "empty\_event" and rally-endings frames with "point".
    \item Step 2 involved revisiting the already marked frames and replacing the temporary tags with their corresponding, detailed annotations.
\end{itemize}

\begin{figure}[h]
    \centering
    \includegraphics[width=1.0\linewidth]{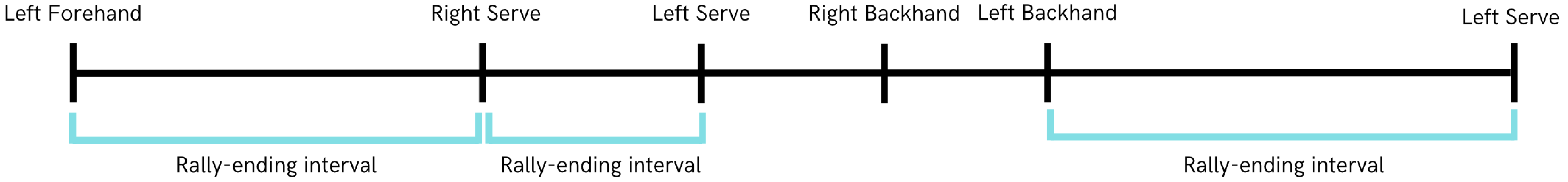}
    \caption{Illustration of the frame interval containing a rally-ending event. The rally-ending will, by definition, occur within these intervals, which substantially reduces the number of frames that need to be inspected.}
    \label{fig:intervals}
\end{figure}

For videos that already include stroke timestamps, Step 1 can be skipped for strokes, as the annotation script uses the provided timestamps to jump directly to the relevant frames. Rally-ending events, however, did not have predefined timestamps in any of the videos. Consequently, both steps were always required for rally-ending annotations. Nevertheless, once strokes had been annotated, identifying rally endings was much simpler, as a rally-ending necessarily occurs after a stroke and before the subsequent serve. This yields frame intervals, which our frame annotation script can jump between. These frame intervals are illustrated in \autoref{fig:intervals}. Despite these methodological advantages, the process required many hours of careful review and repeated verification.

\subsection{Fine-grained Human Posture and Technique During Ball Strikes}
\label{sec:shot_human}
From the above description, it follows that our annotations consist of the following information: stroke information, as well as  body posture and feet position during the stroke.

Hierarchy used in our annotations: \emph{stroke} $\rightarrow$ \emph{lean} $\rightarrow$ \emph{Feet information}.

\subsubsection{Stroke classification}
\begin{figure}[H]
\centering

\begin{subfigure}[b]{\textwidth}
\centering
\begin{subfigure}[b]{0.13\textwidth}
    \includegraphics[width=\linewidth]{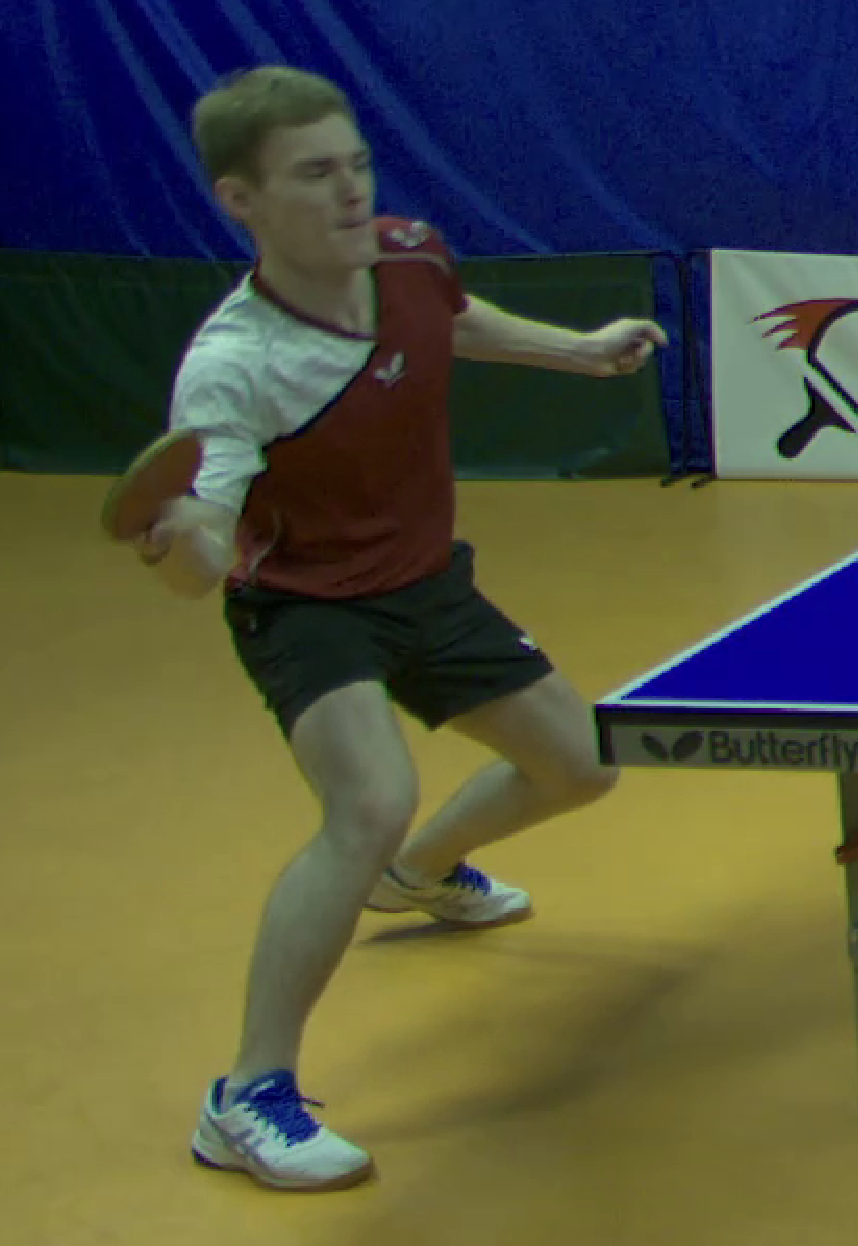}
    \caption{Block}
\end{subfigure}
\hfill
\begin{subfigure}[b]{0.13\textwidth}
    \includegraphics[width=\linewidth]{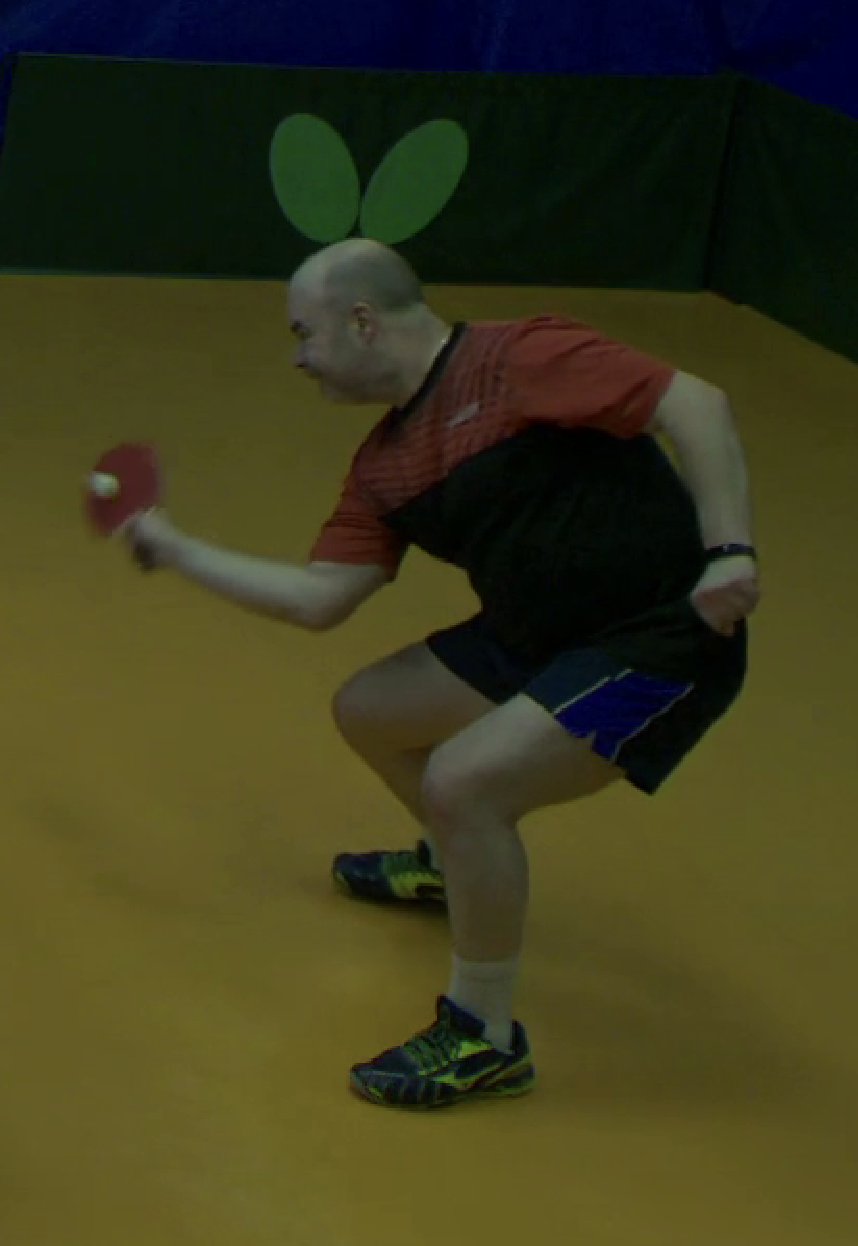}
    \caption{Chop}
\end{subfigure}
\hfill
\begin{subfigure}[b]{0.13\textwidth}
    \includegraphics[width=\linewidth]{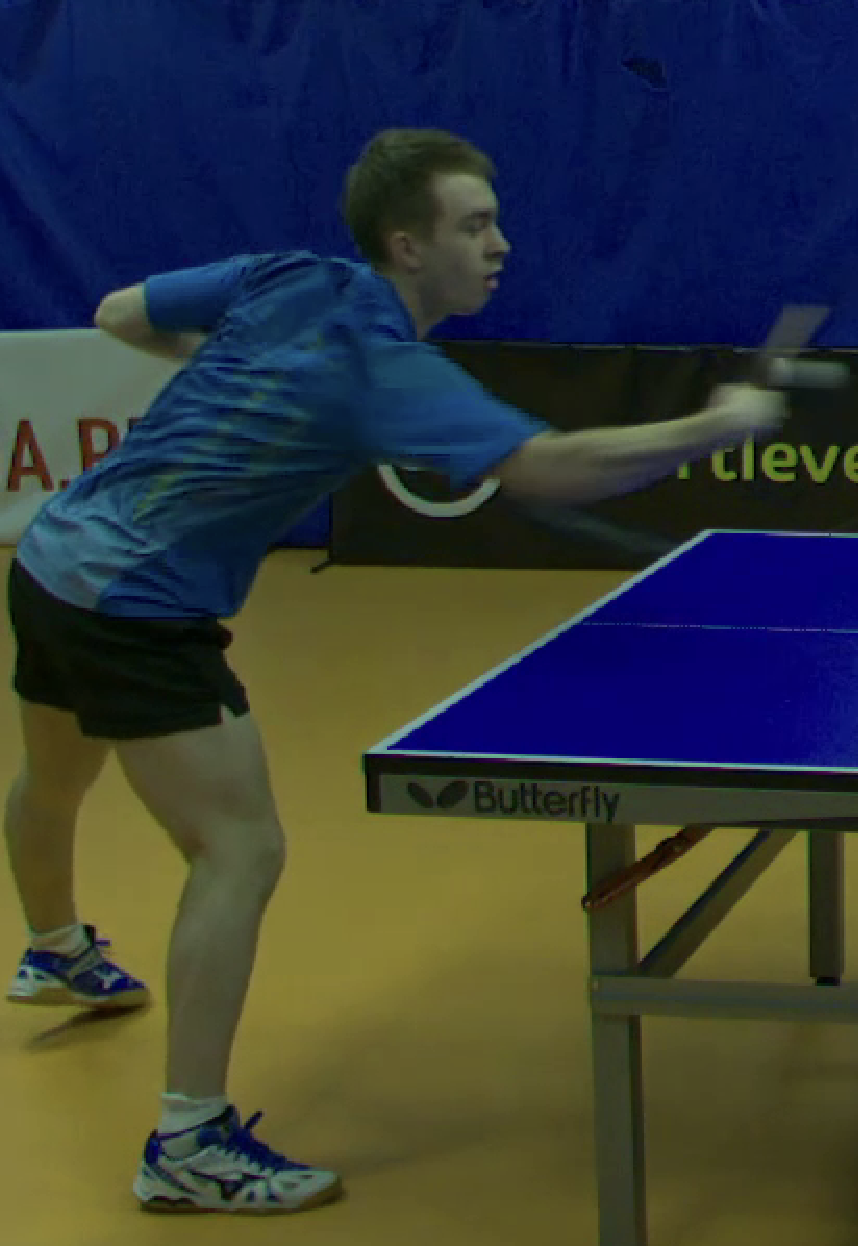}
    \caption{Flick}
\end{subfigure}
\hfill
\begin{subfigure}[b]{0.13\textwidth}
    \includegraphics[width=\linewidth]{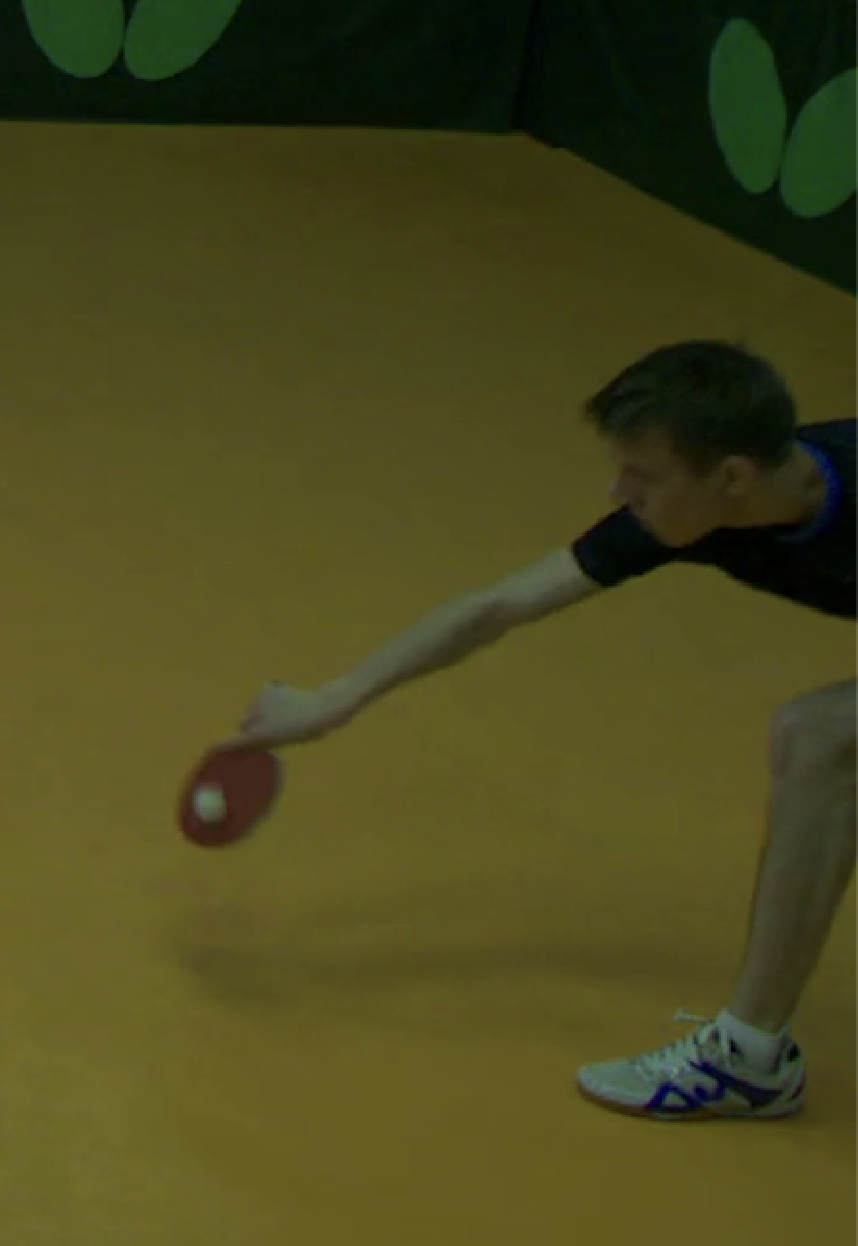}
    \caption{Lob}
\end{subfigure}
\hfill
\begin{subfigure}[b]{0.13\textwidth}
    \includegraphics[width=\linewidth]{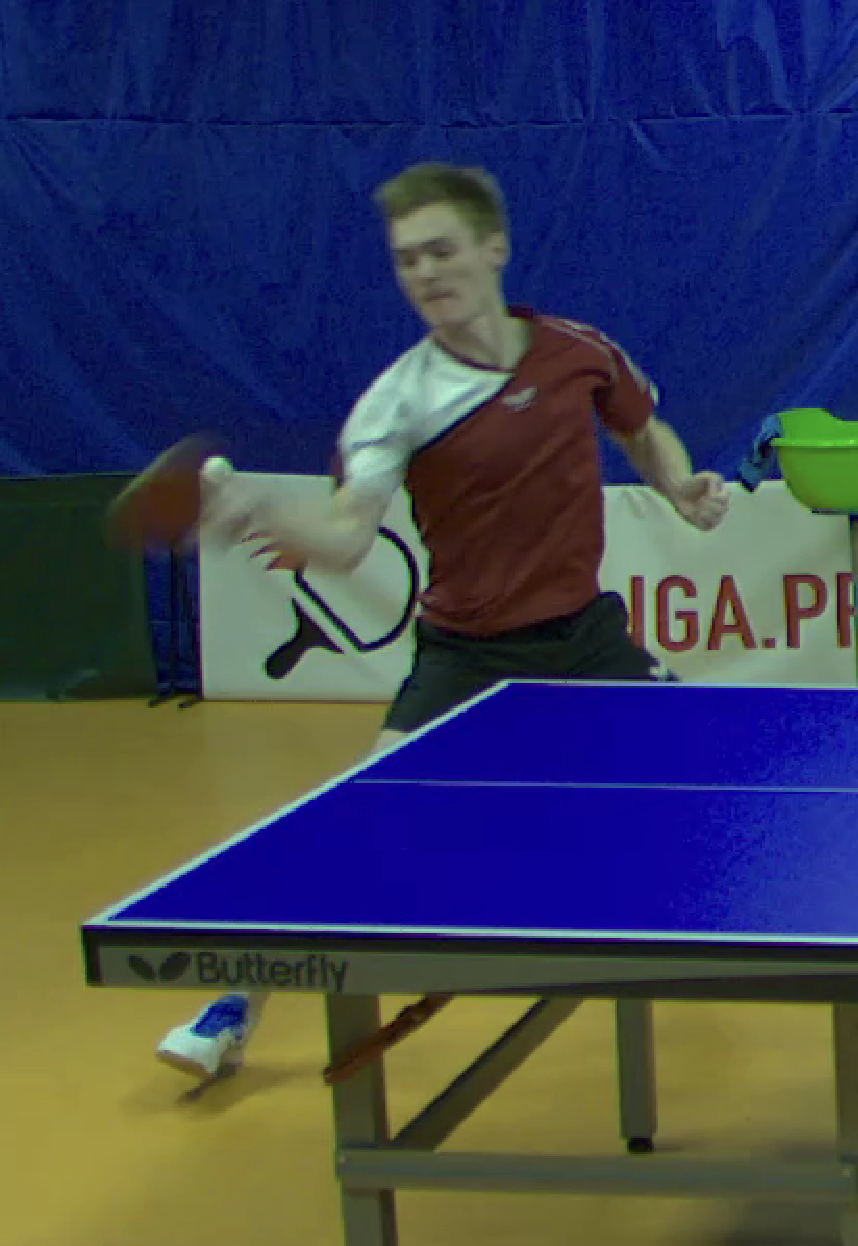}
    \caption{Loop}
\end{subfigure}
\hfill
\begin{subfigure}[b]{0.13\textwidth}
    \includegraphics[width=\linewidth]{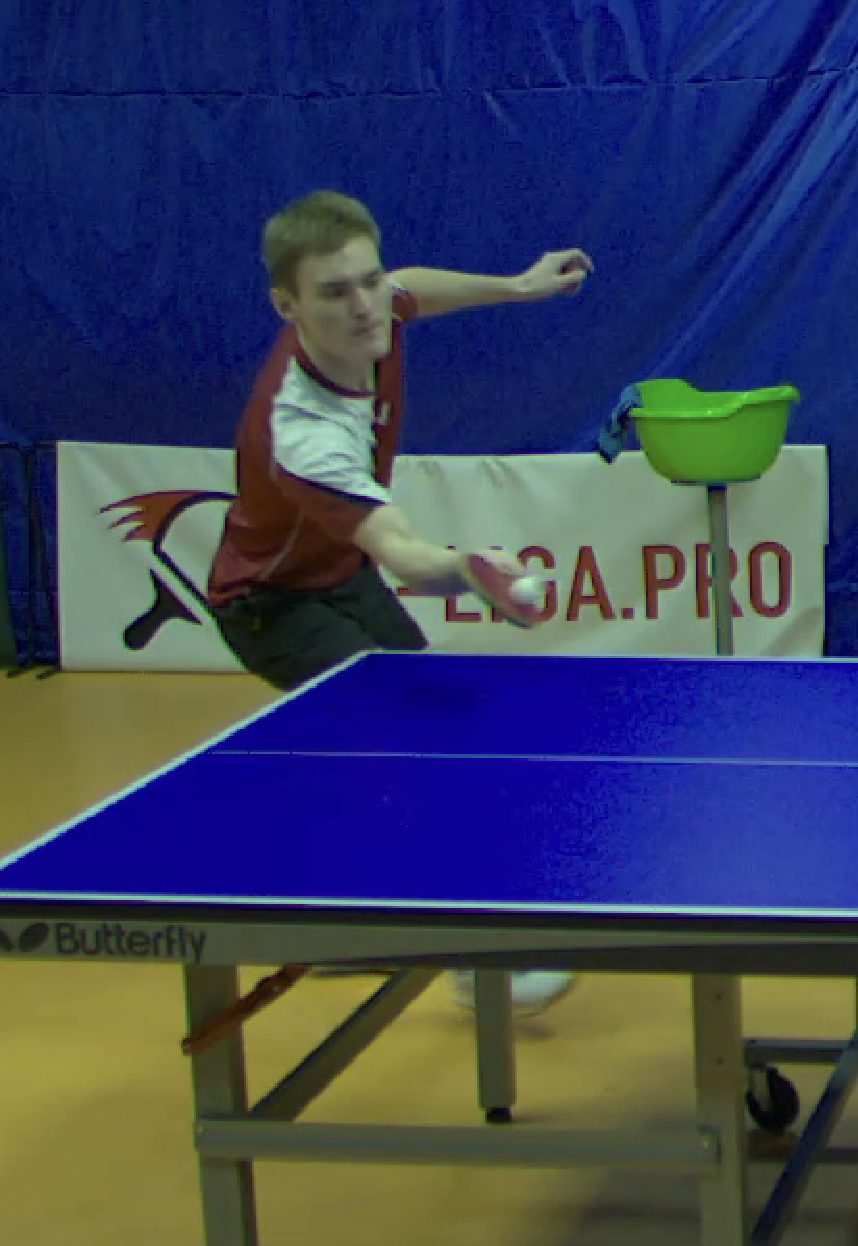}
    \caption{Push}
\end{subfigure}
\hfill
\begin{subfigure}[b]{0.13\textwidth}
    \includegraphics[width=\linewidth]{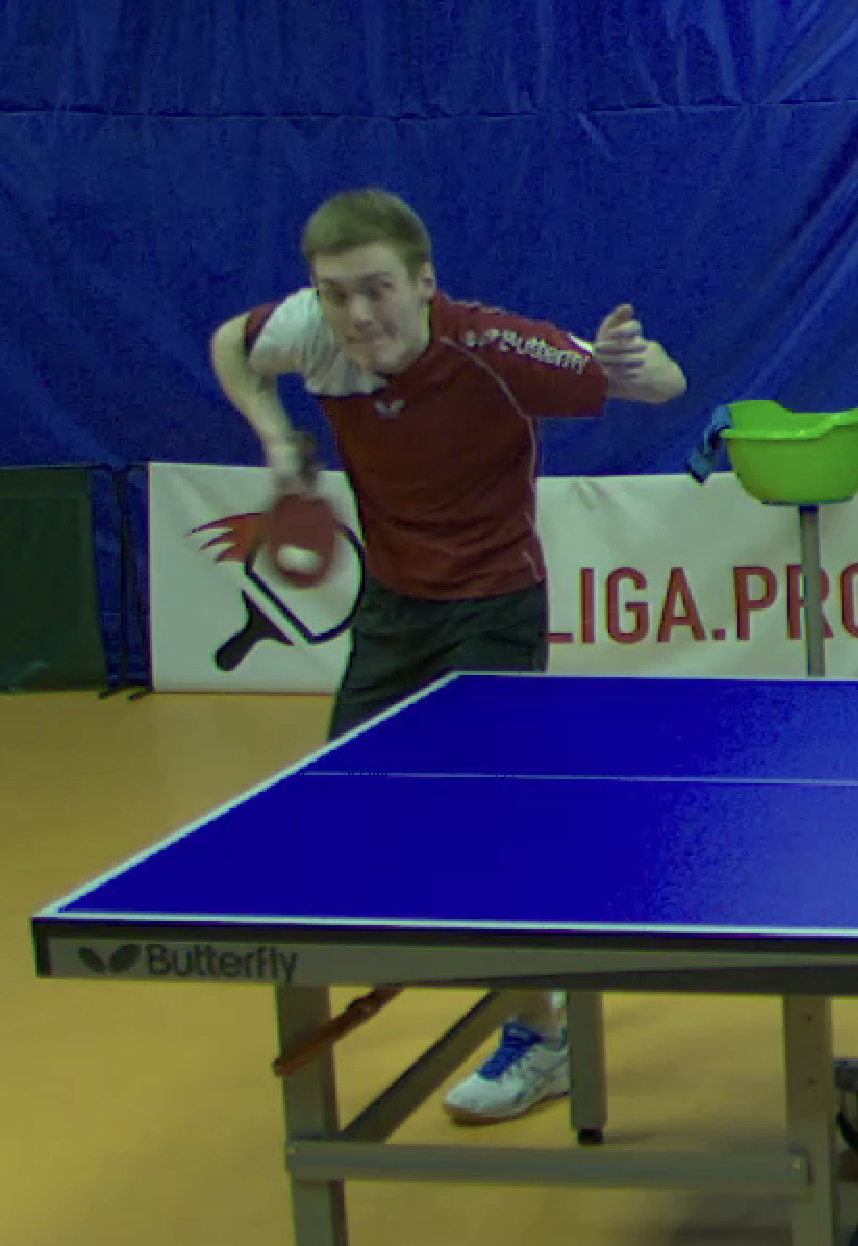}
    \caption{Serve}
\end{subfigure}
\end{subfigure}

\vspace{2em}

\begin{subfigure}[b]{\textwidth}
\centering
\begin{subfigure}[b]{0.18\textwidth}
    \includegraphics[width=\linewidth]{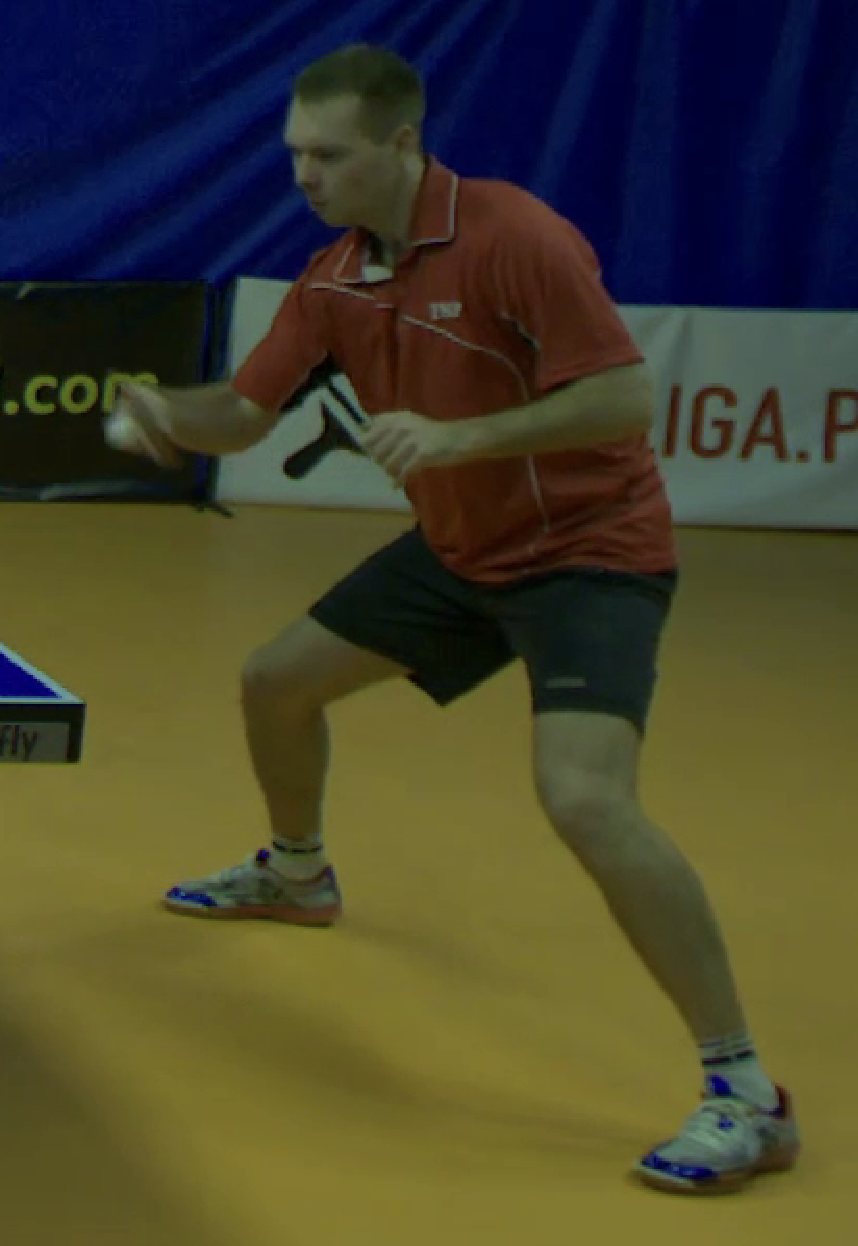}
    \caption{Neutral}
\end{subfigure}
\hfill
\begin{subfigure}[b]{0.18\textwidth}
    \includegraphics[width=\linewidth]{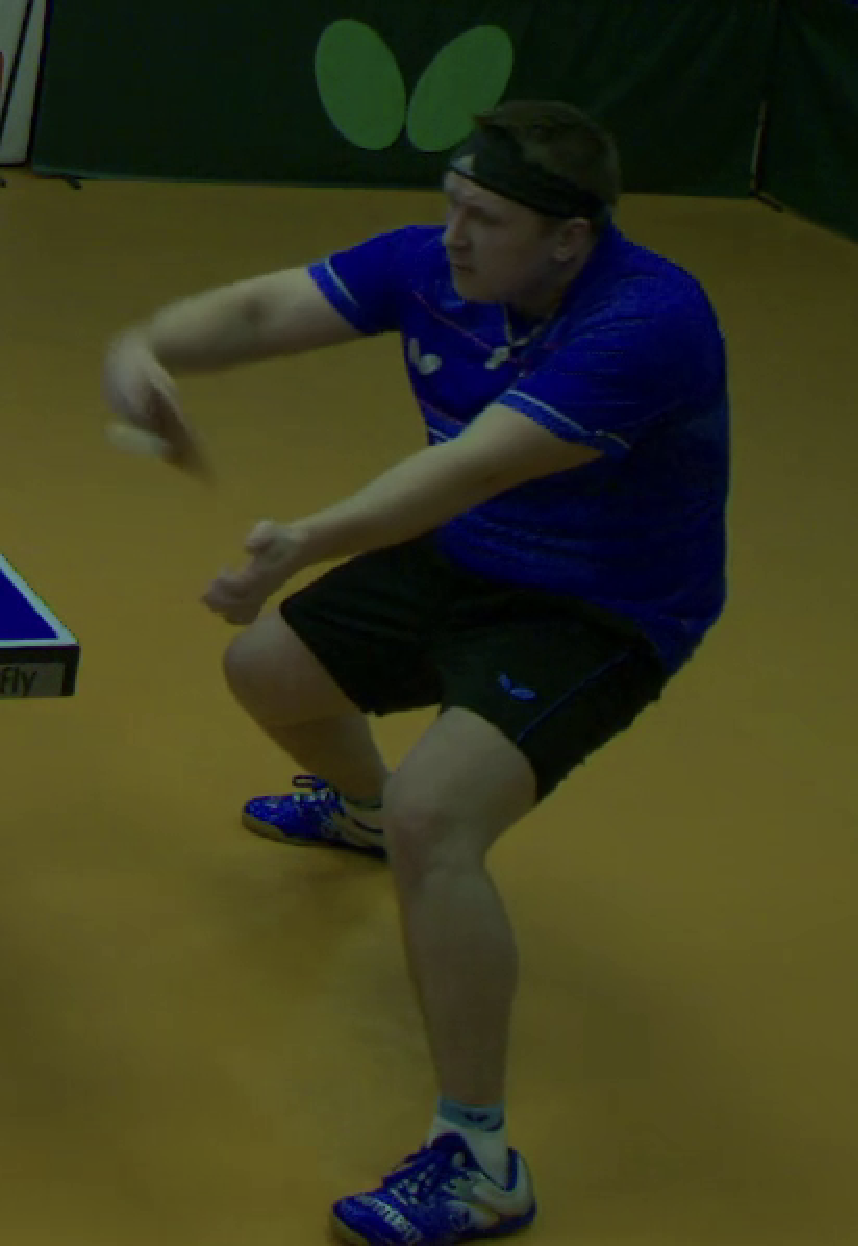}
    \caption{Back-heavy}
\end{subfigure}
\hfill
\begin{subfigure}[b]{0.18\textwidth}
    \includegraphics[width=\linewidth]{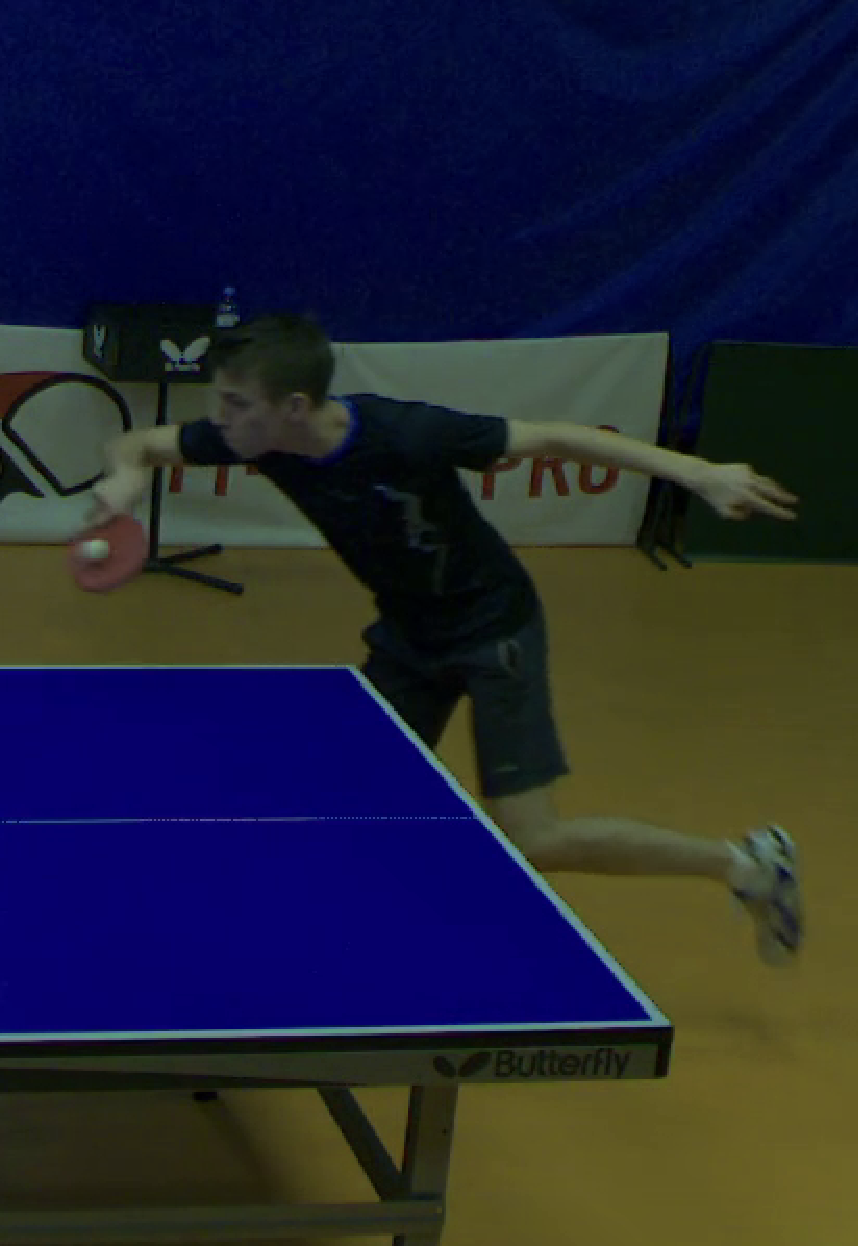}
    \caption{Front-heavy}
\end{subfigure}
\hfill
\begin{subfigure}[b]{0.18\textwidth}
    \includegraphics[width=\linewidth]{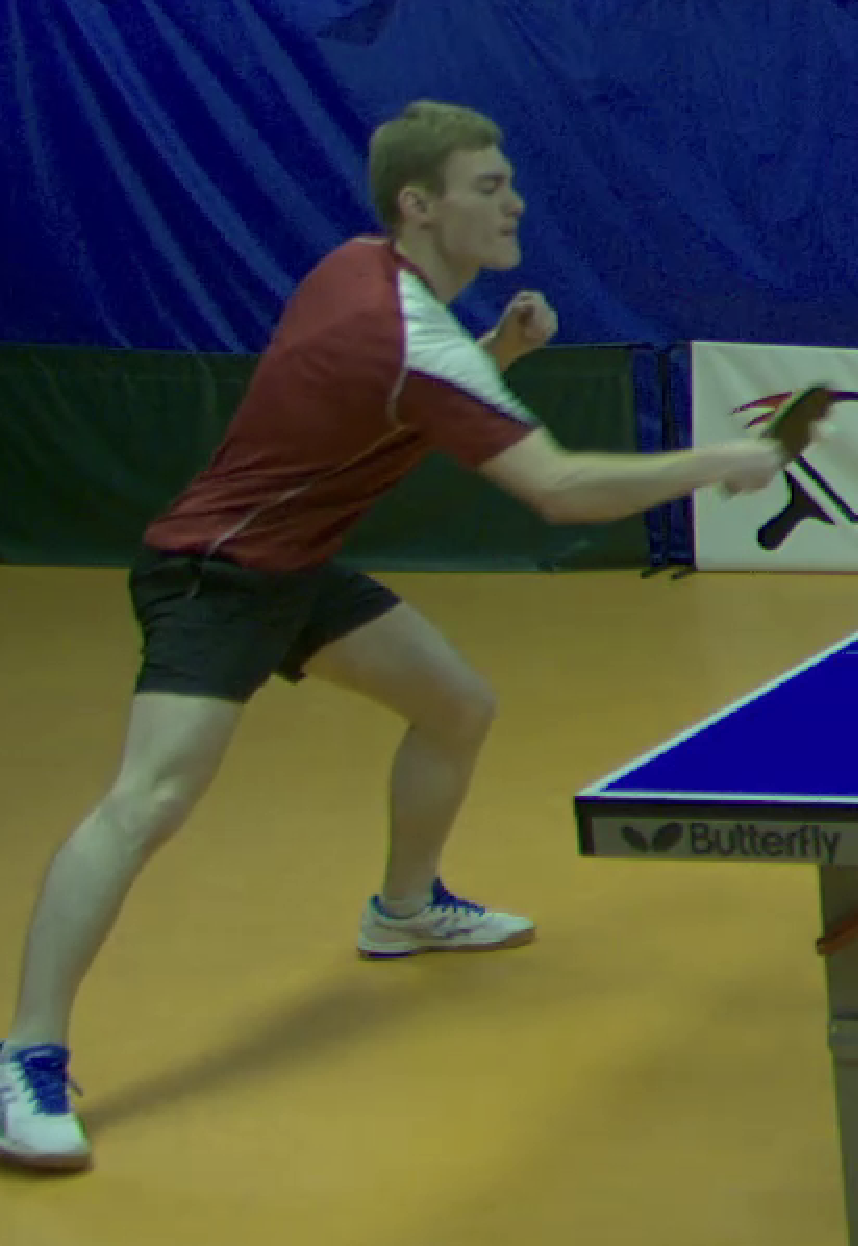}
    \caption{Left-leaning}
\end{subfigure}
\hfill
\begin{subfigure}[b]{0.18\textwidth}
    \includegraphics[width=\linewidth]{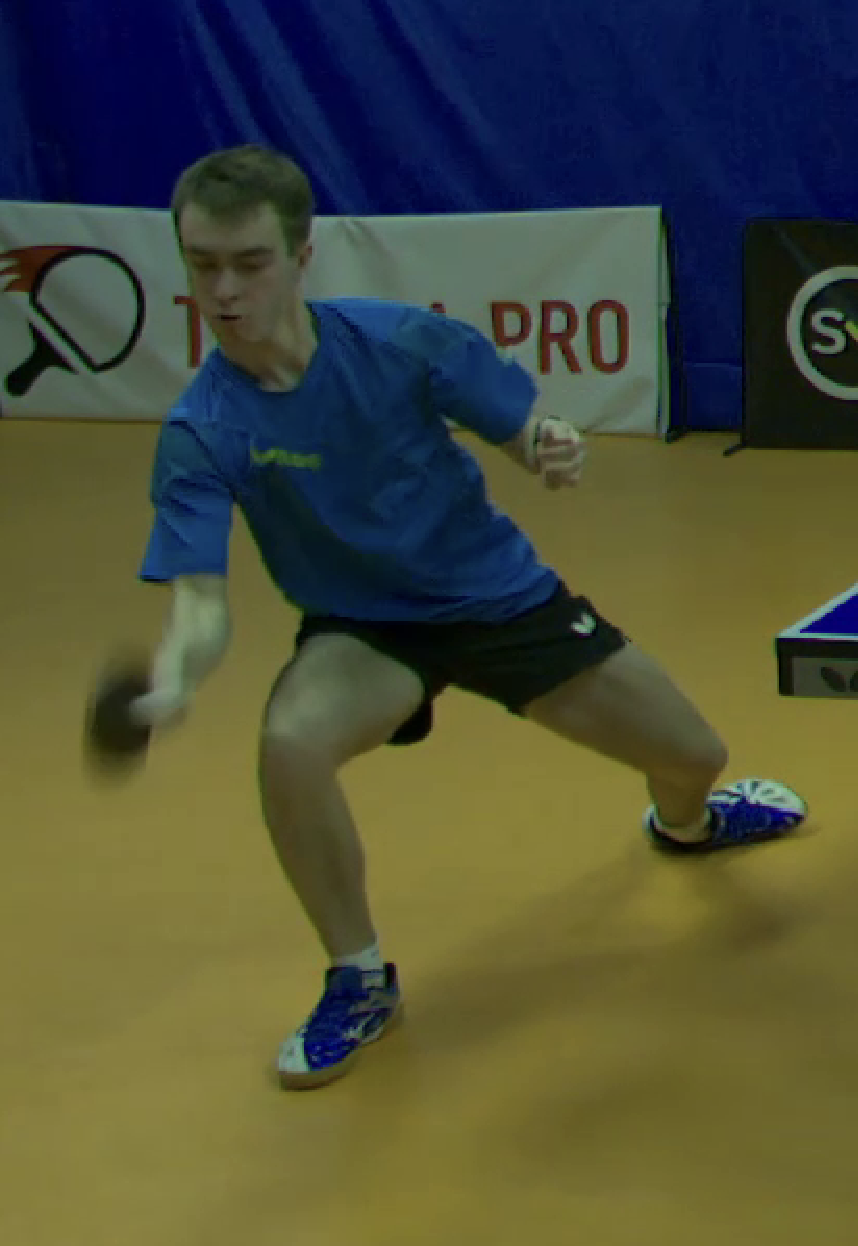}
    \caption{Right-leaning}
\end{subfigure}
\end{subfigure}

\caption{Examples of stroke types (top row) and lean types (bottom row).}
\label{fig:strokes_lean_combined}
\end{figure}

We define a structured taxonomy for classifying the different stroke types. Initially, the strokes are divided into two main classes, "left" or "right", depending on which side of the table the stroke was executed on, from the camera's point of view. Secondly, the strokes are categorized into two high-level classes: forehand and backhand. These top-level categories comprise more specific stroke techniques, such as loop, push, and flick. These types are well known within the table tennis community, and capture subtle differences between strokes that resemble each other, such as loop and smash. All the specific stroke techniques are shown in the top row of \autoref{fig:strokes_lean_combined}. 


The final stroke label concatenates all three levels of the hierarchy using underscores, for example: "left\_forehand\_serve". The exact label counts can be seen in \autoref{sec:stroke_counts}.

Deciding on what frames constitute a stroke was non-trivial. We decided to only include the exact frame of the ball-racket impact. This was despite the fact that a stroke arguably consists of a frame sequence covering the preparatory movements before impact, all the way to the follow-through after impact, as opposed to the bounce and net, which are by definition instantaneous events. On the other hand, deciding the exact boundaries of the frame sequence can be fuzzy and difficult to agree on. Going further into this problem is therefore left for future work. \\ Another consequence of this decision is that the moment of impact is not always captured in the 120 fps videos. In these situations, either the preceding or the subsequent frame could be selected. As the OpenTTGames dataset uses the subsequent frame, we adhere to the same choice. By only using the frame of ball-racket impact, the annotation process is made significantly easier.

\subsubsection{Lean information - posture classification}
\begin{figure}[H]
    \centering
    \includegraphics[width=0.2\linewidth]{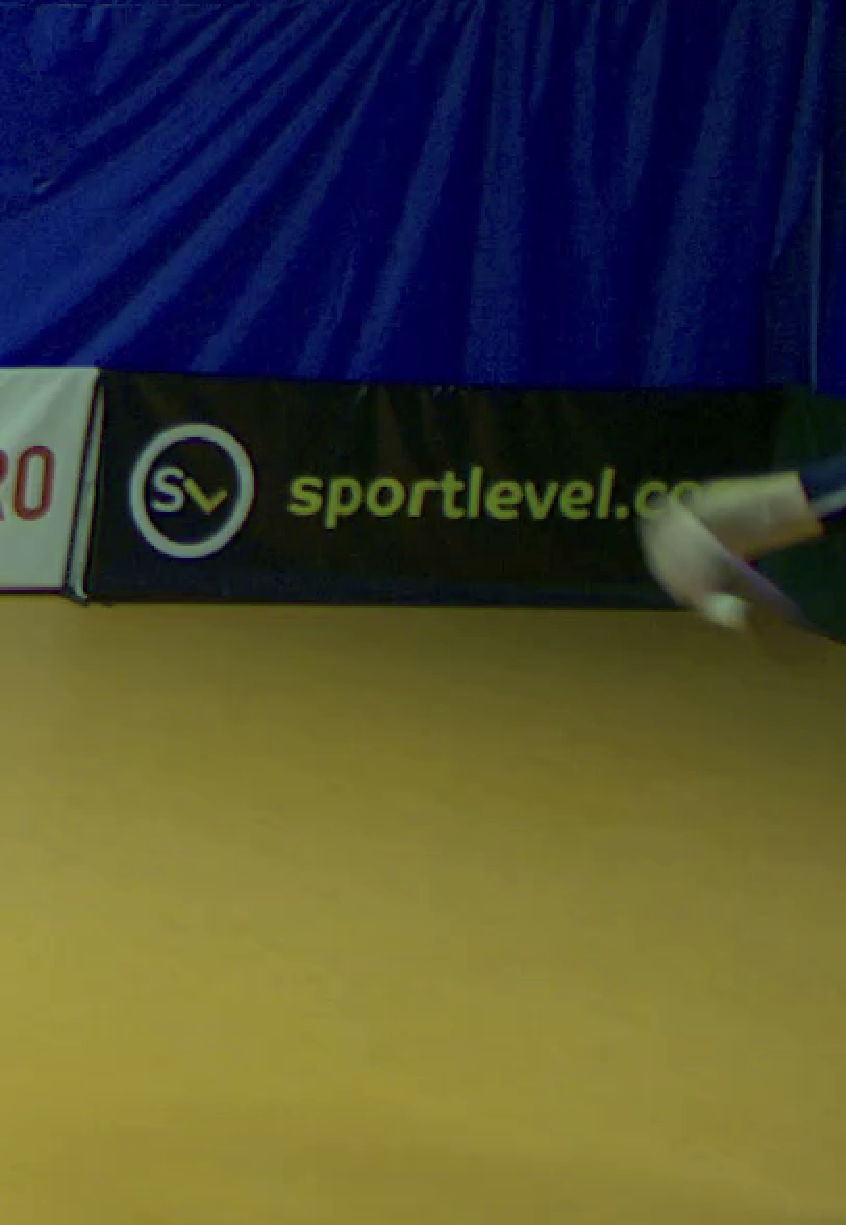}
    \caption{A stroke in which the player is nearly fully outside the camera frame. The stroke is annotated as "right\_backhand\_loop unknown unknown".}
    \label{fig:outside_frame}
\end{figure}

The second component of the label represents the lean information, which is posture classification. The bottom row of \autoref{fig:strokes_lean_combined} illustrates all lean labels used, excluding the “unknown” label. When determining if a player is off balance, only the frame of ball-racket contact is considered. Here, it was only checked if the striking player's center of gravity was off in one of the four directions or neutral. However, the players can be off balance between two of the directions. In these few cases, the example had to be annotated as the side the player leaned towards the most, even if it was very slight. Future work could include adding leans between the existing ones to get a finer-grained annotation. \autoref{fig:outside_frame} shows one of the rare cases where the lean is also unknown during a stroke. Here, only the racket and ball are visible during ball-racket impact, justifying a stroke annotation.

\subsubsection{Feet information}
\begin{figure}[H]
\centering
\begin{subfigure}[t]{0.20\textwidth}
    \includegraphics[width=\linewidth]{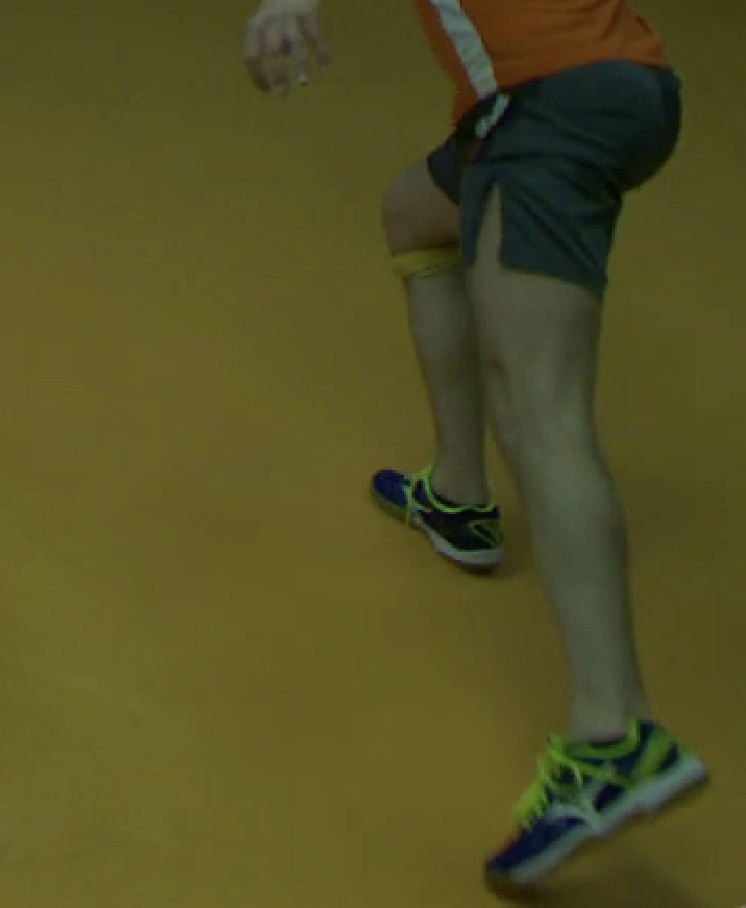}
    \caption{Left foot lifted}
    \label{fig:lifted}
\end{subfigure}
\hfill
\begin{subfigure}[t]{0.20\textwidth}
    \includegraphics[width=\linewidth]{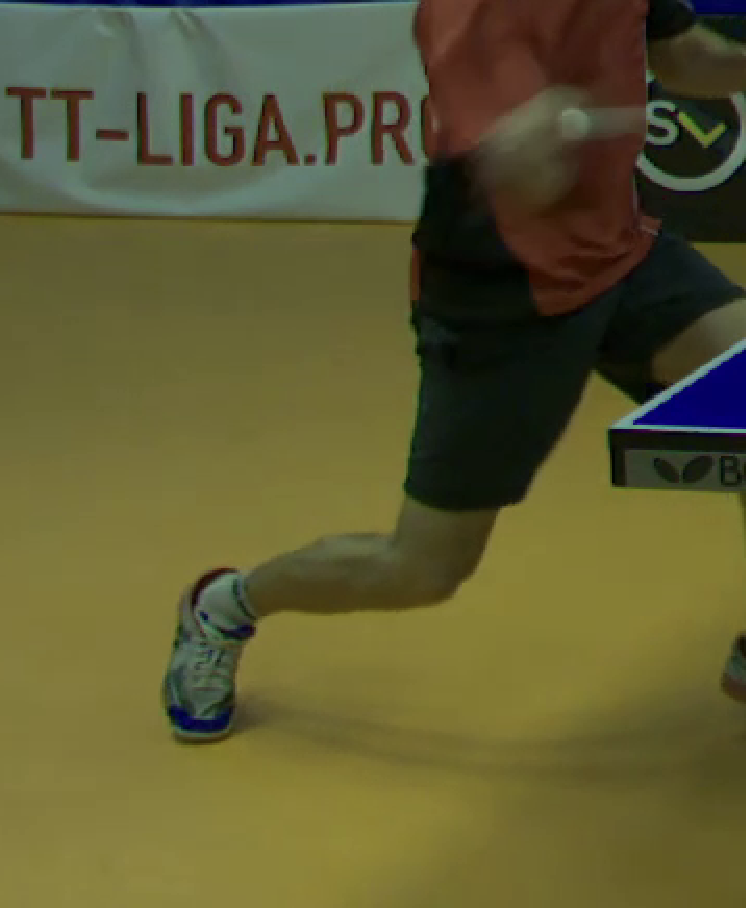}
    \caption{\raggedright Right foot planted}
    \label{fig:planted}
\end{subfigure}
\hfill
\begin{subfigure}[t]{0.20\textwidth}
    \includegraphics[width=\linewidth]{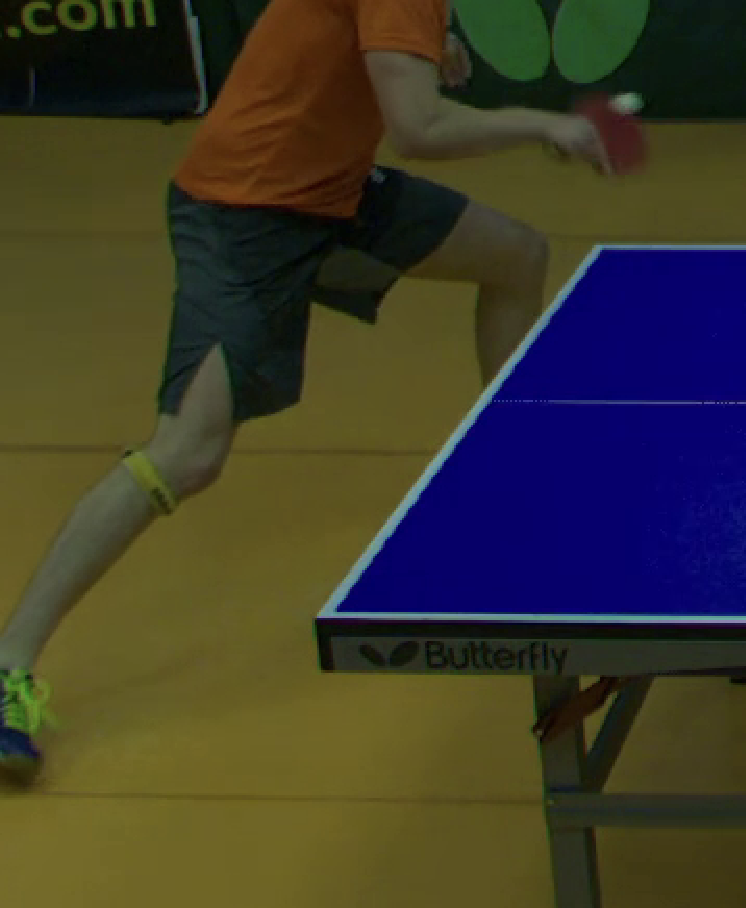}
    \caption{\raggedright Left foot inferred to be planted}
    \label{fig:inferred}
\end{subfigure}
\hfill
\begin{subfigure}[t]{0.20\textwidth}
    \includegraphics[width=\linewidth]{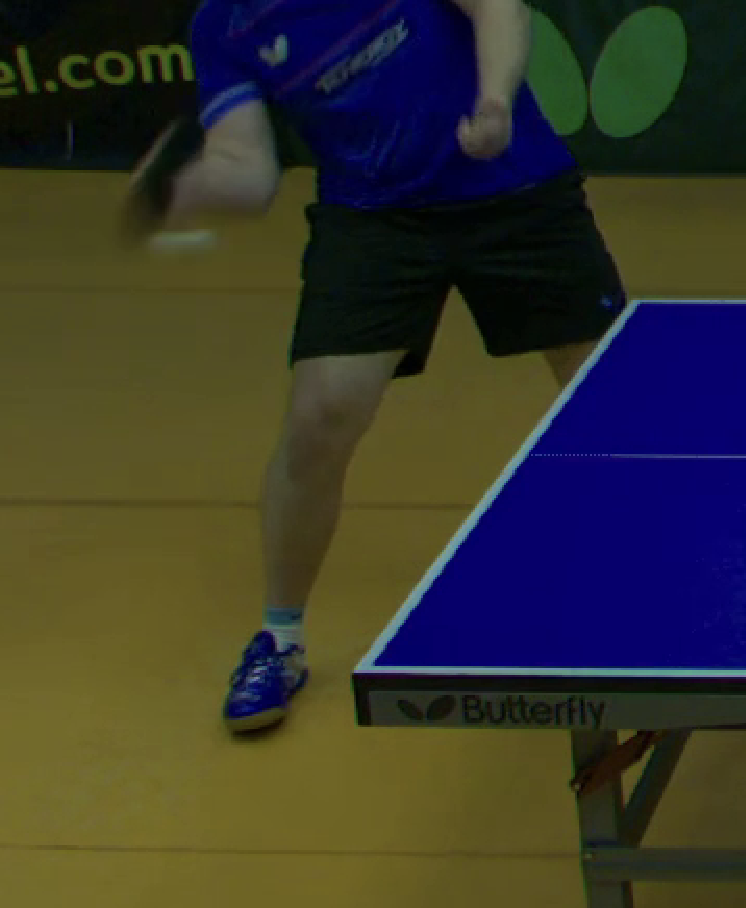}
    \caption{Unknown}
    \label{fig:unknown}
\end{subfigure}
\caption{Examples of non-trivial foot-position configurations together with their counter-examples, demonstrating how the foot-position classification rules are applied.}
\label{fig:strokes_leg_information}
\end{figure}

The final stage of the stroke annotations was the leg information. We analyze each foot separately, assessing whether it was lifted during the stroke. Annotating foot information was trivial for many examples; however, some additional rules had to be applied for the non-trivial cases. Some non-trivial examples are shown in \autoref{fig:strokes_leg_information}.

We decided that slight contact with the ground does not constitute the planted label. The same is the case when the foot is being dragged across the floor. In \autoref{fig:lifted}, the left foot makes only marginal contact, limited to the shoe’s edge, and is thus annotated as lifted. Conversely, in \autoref{fig:planted}, the right foot’s toes are pressed to the ground, and the foot is therefore considered planted.

Another important distinction is to be found between \autoref{fig:inferred} and \autoref{fig:unknown}. In both instances, the player's left leg is occluded, but in the first case, the leg can be inferred to be planted by considering the rest of the body. In the latter case, the leg was slightly moving, making it impossible to tell whether the foot was slightly lifted or not, constituting the "unknown" label.




\begin{table}[H]
\centering
\caption{Counts of annotated human posture and technique labels across training videos.}
\begin{tabular}{lcccccc}
\toprule
\textbf{Label} & \textbf{Train 1} & \textbf{Train 2} & \textbf{Train 3} & \textbf{Train 4} & \textbf{Train 5} & \textbf{Total} \\ 
\midrule
\multicolumn{7}{l}{\textbf{Strokes}} \\[2pt]
Blocks               & 51    & 42    & 27    & 15    & 6     & 141 \\
Chops                & 0     & 2     & 7     & 20    & 0     & 29 \\
Flicks               & 6     & 12    & 0     & 9     & 14    & 41 \\
Lobs                 & 1     & 4     & 1     & 2     & 0     & 8 \\
Loops                & 63    & 179   & 65    & 42    & 82    & 431 \\
Pushes                & 13    & 64    & 23    & 53    & 94    & 247 \\
Serves               & 24    & 92    & 30    & 30    & 52    & 228 \\
Smashes               & 3     & 4     & 0     & 2     & 0     & 9 \\
\hline
\textbf{Total (Strokes)} & \textbf{161} & \textbf{399} & \textbf{153} & \textbf{173} & \textbf{248} & \textbf{1,134} \\[6pt]
\multicolumn{7}{l}{\textbf{Lean Information}} \\[2pt]
Neutral              & 151   & 291   & 87    & 103   & 175   & 807 \\
Back heavy           & 3     & 49    & 37    & 38    & 21    & 148 \\
Front heavy          & 0     & 14    & 4     & 5     & 21    & 44 \\
Right leaning        & 3     & 23    & 16    & 19    & 20    & 81 \\
Left leaning         & 1     & 13    & 6     & 6     & 11    & 37 \\
\hline
\textbf{Total (Lean)} & \textbf{158} & \textbf{390} & \textbf{150} & \textbf{171} & \textbf{248} & \textbf{1,117} \\[6pt]
\multicolumn{7}{l}{\textbf{Leg Information}} \\[2pt]
Both feet lifted     & 5     & 4     & 1     & 1     & 0     & 11 \\
Both feet planted    & 118   & 248   & 84    & 88    & 167   & 705 \\
Left foot lifted     & 16    & 108   & 34    & 38    & 57    & 253 \\
Right foot lifted    & 14    & 8     & 7     & 15    & 12    & 56 \\
\hline
\textbf{Total (Leg)} & \textbf{153} & \textbf{368} & \textbf{126} & \textbf{142} & \textbf{236} & \textbf{1,025} \\
\bottomrule
\end{tabular}
\label{tab:label_count}
\end{table}

\begin{table}[H]
\centering
\caption{Counts of annotated human posture and technique labels across test videos.}
\begin{tabular}{lcccccccc}
\toprule
\textbf{Label} & \textbf{Test 1} & \textbf{Test 2} & \textbf{Test 3} & \textbf{Test 4} & \textbf{Test 5} & \textbf{Test 6} & \textbf{Test 7} & \textbf{Total} \\ 
\midrule
\multicolumn{7}{l}{\textbf{Strokes}} \\[2pt]
Blocks               & 30    & 0     & 0     & 0     & 0     & 7     & 9     & 46 \\
Chops                & 0     & 0     & 0     & 2     & 0     & 0     & 0     & 2 \\
Flicks               & 1     & 0     & 0     & 5     & 3     & 3     & 12    & 24 \\
Lobs                 & 1     & 1     & 0     & 0     & 0     & 0     & 0     & 2 \\
Loops                & 41    & 26    & 14    & 27    & 12    & 17    & 15    & 152 \\
Pushs                & 2     & 0     & 4     & 12    & 5     & 4     & 5     & 32 \\
Serves               & 9     & 2     & 5     & 23    & 7     & 8     & 8     & 62 \\
Smashs               & 0     & 0     & 1     & 2     & 0     & 0     & 0     & 3 \\
\hline
\textbf{Total (Strokes)} & \textbf{84} & \textbf{29} & \textbf{24} & \textbf{71} & \textbf{27} & \textbf{39} & \textbf{49} & \textbf{323} \\[6pt]
\multicolumn{7}{l}{\textbf{Lean Information}} \\[2pt]
Neutral              & 72    & 29    & 16    & 47    & 23    & 26    & 35    & 248 \\
Back heavy           & 7     & 0     & 3     & 10    & 3     & 3     & 6     & 32 \\
Front heavy          & 1     & 0     & 2     & 3     & 0     & 1     & 2     & 9 \\
Right leaning        & 2     & 0     & 1     & 6     & 0     & 5     & 4     & 18 \\
Left leaning         & 2     & 0     & 1     & 2     & 1     & 1     & 1     & 8 \\
\hline
\textbf{Total (Lean)} & \textbf{84} & \textbf{29} & \textbf{23} & \textbf{68} & \textbf{27} & \textbf{36} & \textbf{48} & \textbf{315} \\[6pt]
\multicolumn{7}{l}{\textbf{Leg Information}} \\[2pt]
Both feet lifted             & 0     & 0     & 0     & 9     & 0     & 1     & 3     & 13 \\
Both feet planted              & 65    & 28    & 13    & 28    & 19    & 19    & 19    & 191 \\
Left foot lifted     & 13    & 1     & 3     & 23    & 1     & 10    & 4     & 55 \\
Right foot lifted    & 2     & 0     & 3     & 6     & 6     & 3     & 15    & 35 \\
\hline
\textbf{Total (Leg)} & \textbf{80} & \textbf{29} & \textbf{19} & \textbf{66} & \textbf{26} & \textbf{33} & \textbf{41} & \textbf{294} \\
\bottomrule
\end{tabular}
\label{tab:label_count_test}
\end{table}

\subsubsection{Data summary of the fine-grained taxonomy}
This fine-grained taxonomy results in a large number of highly specific labels, many of which occur only a few times. For this reason, we encourage users of the dataset to aggregate labels according to their task or analysis needs. For instance, one may pool all forehand loops regardless of lean or foot placement, group labels based solely on lean categories, or apply any other form of label consolidation appropriate for a given use case. The counts of the separate parts of the labels are listed in \autoref{tab:label_count} and \autoref{tab:label_count_test}.

\subsection{Fine-grained point outcome}\label{sec:point_end}
\begin{figure}[H]
\centering
\begin{subfigure}[b]{0.15\textwidth}
    \includegraphics[width=\linewidth]{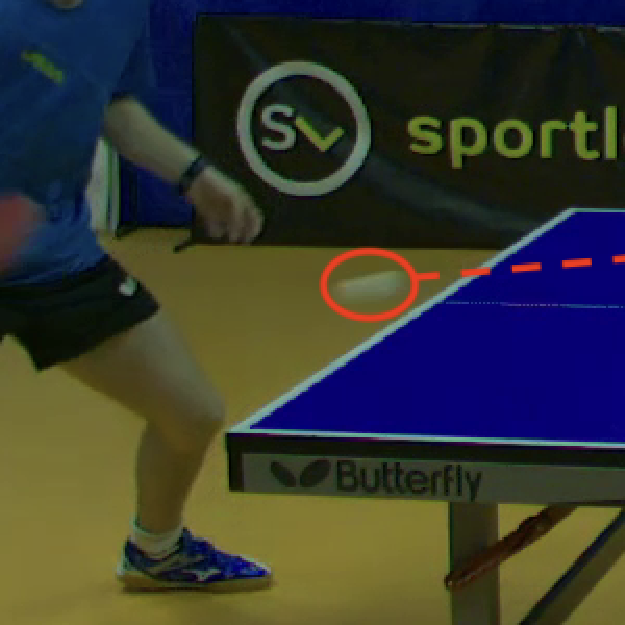}
    \caption{Out}
    \label{fig:out}
\end{subfigure}
\hfill
\begin{subfigure}[b]{0.15\textwidth}
    \includegraphics[width=\linewidth]{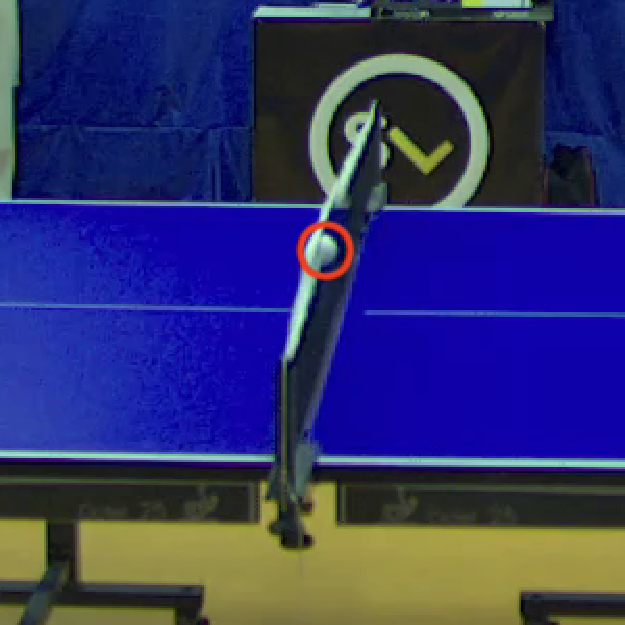}
    \caption{Net}
    \label{fig:net}
\end{subfigure}
\hfill
\begin{subfigure}[b]{0.15\textwidth}
    \includegraphics[width=\linewidth]{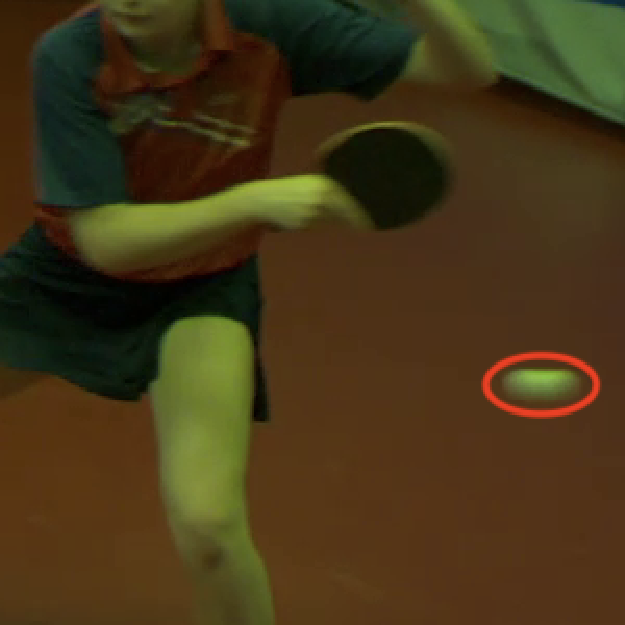}
    \caption{Winner}
    \label{fig:winner}
\end{subfigure}
\hfill
\begin{subfigure}[b]{0.15\textwidth}
    \includegraphics[width=\linewidth]{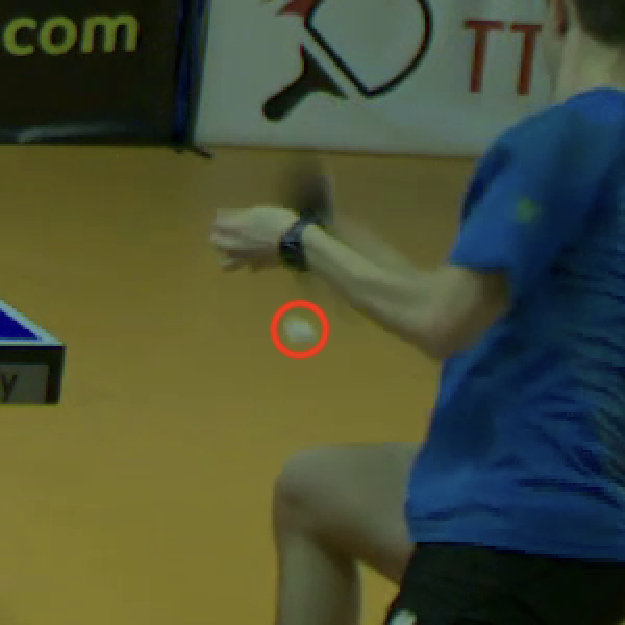}
    \caption{Missing ball}
    \label{fig:missing}
\end{subfigure}
\hfill
\begin{subfigure}[b]{0.15\textwidth}
    \includegraphics[width=\linewidth]{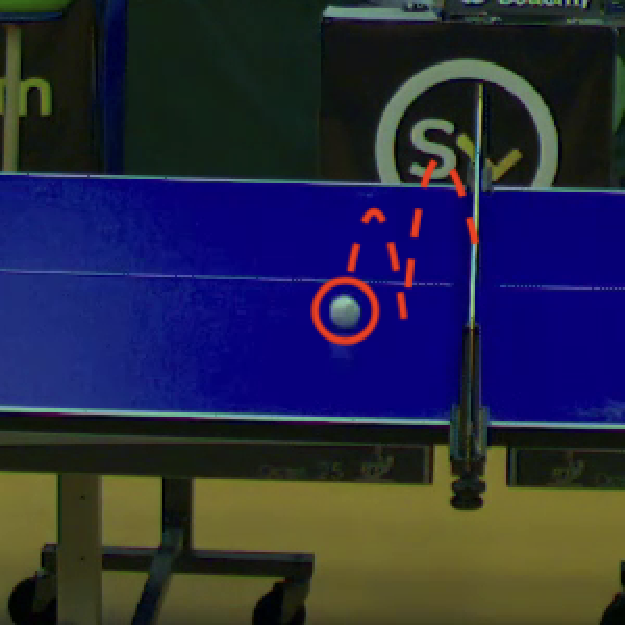}
    \caption{Double bounce}
    \label{fig:double_bounce}
\end{subfigure}
\hfill
\begin{subfigure}[b]{0.15\textwidth}
    \includegraphics[width=\linewidth]{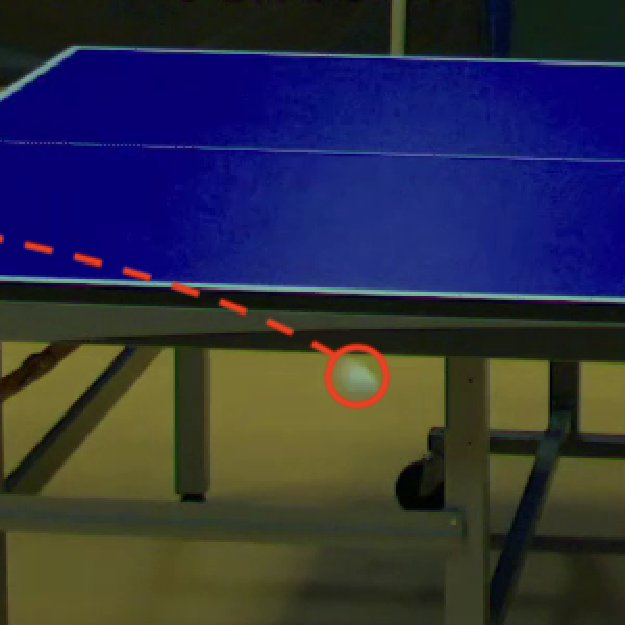}
    \caption{Miss own side}
    \label{fig:miss_on_own_side}
\end{subfigure}
\caption{Representative examples of the rally-ending categories defined in the dataset.}
\label{fig:rally_endings}
\end{figure}

In addition to the above annotations, we also decided to label how the rallies ended. Like for the stroke annotations, the rally outcomes are prefixed with the player who hit the rally-ending ball. The rally endings have been distributed into the following six categories, which should cover the various ways a rally may conclude, providing a detailed and fine-grained distinction between outcomes. Examples of the rally outcomes are shown in \autoref{fig:rally_endings}. The six categories are illustrated below for the left-side player (for the right-side player, the logic is mirrored).

\begin{itemize}
    \item (a) Out-class:  \\
    \textbf{Description}: When the ball has fully passed the table, either below the side edges or past the back line. \\
    \textbf{Prefix}: The player who hits the ball out. \\
    \textbf{Example}: The left player hits the ball out.
    \begin{itemize}
        \item  label: left\_out.
        \item Point awarded to the right player.
    \end{itemize}
    
    \item (b) Net-class:\\
    \textbf{Description}: The ball is stopped by the net or hits the striker’s own side of the table. If the ball touches the net but still crosses it, or if it later goes out, no net-induced rally ending is annotated. \\
    \textbf{Prefix}: The player who hits the ball into the net or their own side of the table. \\
    \textbf{Example}: The left player hits the ball into the net.
        \begin{itemize}
        \item label: left\_net.
        \item Point awarded to the right player.
    \end{itemize}
    
    \item (c) Winner-class\\
    \textbf{Description}: The ball becomes unreachable for the receiving player, including cases where the ball hits the player’s body. \\
    \textbf{Prefix}: The player who hits the winner. \\
    \textbf{Example}: The left player hits a winner; the right player cannot reach the ball.  \begin{itemize}
        \item label: left\_winner.
        \item Point awarded to the left player.
    \end{itemize}

    \item (d) Not\_hitting\_ball-class: \\ 
    \textbf{Description}: The striker swings for a reachable ball but misses i.e., the ball passes their racket. \\
    \textbf{Prefix}: The player who attempts the hit and misses. \\
    \textbf{Example}: The left player swings and misses.
    \begin{itemize}
        \item label: left\_not\_hitting\_ball.
        \item Point awarded to the right player.
    \end{itemize}
    
    \item (e) Double\_bounce-class:  \\
    \textbf{Description}: The ball bounces twice on one player’s side of the table. \\
    \textbf{Prefix}: The player who hit the ball before the double bounce. \\
    \textbf{Example}: The ball bounces twice on the left player's side.
        \begin{itemize}
        \item label: right\_double\_bounce.
        \item Point awarded to the right player.
    \end{itemize}
    
    \item (f) Miss\_on\_own\_side-class:\\ 
    \textbf{Description}: The ball drops completely below the table on the striker’s own side. If the ball hits the racket edge and flies off to the side, the rally ending is labeled a few frames after contact. \\
    \textbf{Prefix}: The player who hits the ball off their own side. \\
    \textbf{Example}: The left player hits the ball, and it exits the table on their own side.
    \begin{itemize}
        \item label: left\_miss\_on\_own\_side.
        \item Point awarded to the right player.
    \end{itemize}
\end{itemize}


Since each rally-ending label consists of a player-specific prefix and an outcome category, users may, as with the stroke-level labels, consolidate or ignore components of the label depending on their specific use case. Counts of the different rally-ending outcomes are reported in \autoref{tab:point-outcomes-summary} and \autoref{tab:point-outcomes-summary_test} for the training and test videos, respectively.

\begin{table}[H]
\centering
\caption{Counts of annotated rally-endings across training videos.}
\begin{tabular}{lcccccc}
\toprule
\textbf{Label} & \textbf{Train 1} & \textbf{Train 2} & \textbf{Train 3} & \textbf{Train 4} & \textbf{Train 5} & \textbf{Total} \\ 
\midrule
Out                 & 12    & 39    & 20    & 19    & 19    & 109 \\
Net                 & 5     & 28    & 7     & 7     & 18    & 65 \\
Winner              & 4     & 13    & 2     & 3     & 6     & 28 \\
Missing ball - not hitting ball    & 2     & 2     & 1     & 1     & 2     & 8 \\
Double bounce       & 1     & 2     & 0     & 0     & 1     & 4 \\
Miss on own side    & 1     & 3     & 1     & 0     & 4     & 9 \\
\hline
\textbf{Total } & \textbf{25} & \textbf{87} & \textbf{31} & \textbf{30} & \textbf{50} & \textbf{223} \\
\bottomrule
\end{tabular}
\label{tab:point-outcomes-summary}
\end{table}

\begin{table}[H]
\centering
\caption{Counts of annotated rally-endings across test videos.}
\begin{tabular}{lcccccccc}
\toprule
\textbf{Label} & \textbf{Test 1} & \textbf{Test 2} & \textbf{Test 3} & \textbf{Test 4} & \textbf{Test 5} & \textbf{Test 6} & \textbf{Test 7} & \textbf{Total} \\ 
\midrule
Out                 & 4     & 1     & 1     & 13    & 3     & 5     & 4     & 31 \\
Net                 & 1     & 1     & 3     & 7     & 1     & 2     & 3     & 18 \\
Winner              & 0     & 0     & 1     & 0     & 3     & 1     & 1     & 6 \\
Missing ball - not hitting ball   & 2     & 0     & 0     & 0     & 0     & 0     & 0     & 2 \\
Double bounce       & 0     & 0     & 0     & 0     & 0     & 0     & 0     & 0 \\
Miss on own side    & 0     & 0     & 0     & 1     & 0     & 0     & 0     & 1 \\
\hline
\textbf{Total } & \textbf{7} & \textbf{2} & \textbf{5} & \textbf{21} & \textbf{7} & \textbf{8} & \textbf{8} & \textbf{58} \\
\bottomrule
\end{tabular}
\label{tab:point-outcomes-summary_test}
\end{table}

\section{Conclusion}
We presented an extended version of OpenTTGames with frame-accurate annotations of (i) left/right player and stroke type classes, (ii) player posture when the ball have contact with the bat (body lean and legs) 
and (iii) point outcome of the rally.
Hereby we provide a table tennis stroke data set with detailed annotations and a transparent license. By providing structured, frame-accurate stroke, posture, and outcome labels, our dataset enables supervision beyond event spotting and supports learning higher-level tactical patterns in table tennis play. All scripts for inspection and annotation are openly available, facilitating reproducibility and further extensions of the data.




\newpage
\bibliographystyle{unsrt}  
\bibliography{references}  

\newpage
\appendix

\section{Annotation Shorthand and Label Codes}
\label{sec:codes}
The script asks the annotator to enter the stroke category, body-lean details, and leg details by using the designated keywords corresponding to the codes listed in \autoref{tab:stroke-types}, \autoref{tab:lean}, and \autoref{tab:leg}.
\begin{table}[H]
\centering
\caption{Overview the stroke-labels and their respective codes used to annotate stroke frames.}
\begin{tabular}{| l l | l l |}
    \hline
    \textbf{Code} & \textbf{Label} & \textbf{Code} & \textbf{Label} \\
    \hline
    e & empty\_event &  &  \\
    \hline
    lfsv & left\_forehand\_serve & rfsv & right\_forehand\_serve \\
    lbsv & left\_backhand\_serve & rbsv & right\_backhand\_serve \\
    \hline
    lfl & left\_forehand\_loop & rfl & right\_forehand\_loop \\
    lbl & left\_backhand\_loop & rbl & right\_backhand\_loop \\
    \hline
    lfsh & left\_forehand\_chop & rfsh & right\_forehand\_chop \\
    lbsh & left\_backhand\_chop & rbsh & right\_backhand\_chop \\
    \hline
    lfb & left\_forehand\_block & rfb & right\_forehand\_block \\
    lbb & left\_backhand\_block & rbb & right\_backhand\_block \\
    \hline
    lfp & left\_forehand\_push & rfp & right\_forehand\_push \\
    lbp & left\_backhand\_push & rbp & right\_backhand\_push \\
    \hline
    lff & left\_forehand\_flick & rff & right\_forehand\_flick \\
    lbf & left\_backhand\_flick & rbf & right\_backhand\_flick \\
    \hline
    lfs & left\_forehand\_smash & rfs & right\_forehand\_smash \\
    lbs & left\_backhand\_smash & rbs & right\_backhand\_smash \\
    \hline
    lflo & left\_forehand\_lob & rflo & right\_forehand\_lob \\
    lblo & left\_backhand\_lob & rblo & right\_backhand\_lob \\
    \hline
\end{tabular}
\label{tab:stroke-types}
\end{table}

\begin{table}[H]
\centering
\caption{Overview the lean-labels and their respective codes used to annotate stroke frames.}
\begin{tabular}{| l l | l l |}
    \hline
    \textbf{Code} & \textbf{Label} & \textbf{Code} & \textbf{Label} \\
    \hline
    b & back\_heavy & f & front\_heavy \\
    r & right\_leaning & l & left\_leaning \\
    n & neutral & u & unknown \\
    \hline
\end{tabular}
\label{tab:lean}
\end{table}

\begin{table}[H]
\centering
\caption{Overview the leg information-labels and their respective codes used to annotate stroke frames.}
\begin{tabular}{| l l | l l |}
    \hline
    \textbf{Code} & \textbf{Label} & \textbf{Code} & \textbf{Label} \\
    \hline
    b & both\_feet\_planted & bl & both\_feet\_lifted \\
    r & right\_foot\_lifted & l & left\_foot\_lifted \\
    u & unknown &  &  \\
    \hline
\end{tabular}
\label{tab:leg}
\end{table}

\begin{table}[H]
\centering
\caption{Overview the point outcome-labels and their respective codes used to annotate rally-ending frames.}
\begin{tabular}{| l l | l l |}
    \hline
    \textbf{Code} & \textbf{Label} & \textbf{Code} & \textbf{Label} \\
    \hline
    ln & left\_net & rn & right\_net \\
    lnb & left\_not\_hitting\_ball & rnb & right\_not\_hitting\_ball \\
    lw & left\_winner & rw & right\_winner \\
    ld & left\_double\_bounce & rd & right\_double\_bounce \\
    lo & left\_out & ro & right\_out \\
    lm & left\_miss\_on\_own\_side & rm & right\_miss\_on\_own\_side \\
    \hline
\end{tabular}
\label{tab:leg}
\end{table}
\newpage


\section{Stroke counts}\label{sec:stroke_counts}
All the stroke label are concatenated and shown for the training video in \autoref{tab:apendix_train} and test video in \autoref{tab:apendix_test}.

\begin{table}[h]
\centering
\caption{Strokes in the training videos}
\begin{tabular}{lcccccc}
\toprule
\textbf{Stroke} & \textbf{Train 1} & \textbf{Train 2} & \textbf{Train 3} & \textbf{Train 4} & \textbf{Train 5} & \textbf{Total} \\ 
\midrule
left\_forehand\_block     & 7 & 8 & 2 & 1 & 1 & 19 \\
left\_backhand\_block     & 21 & 11 & 8 & 4 & 3 & 47 \\
right\_forehand\_block    & 12 & 10 & 5 & 4 & 1 & 32 \\
right\_backhand\_block    & 11 & 13 & 12 & 6 & 1 & 43 \\
\hline
left\_forehand\_chop      & 0 & 0 & 0 & 2 & 0 & 2 \\
left\_backhand\_chop      & 0 & 0 & 4 & 8 & 0 & 12 \\
right\_forehand\_chop     & 0 & 0 & 0 & 4 & 0 & 4 \\
right\_backhand\_chop     & 0 & 2 & 2 & 6 & 0 & 10 \\
\hline
left\_forehand\_flick     & 2 & 0 & 0 & 0 & 0 & 2 \\
left\_backhand\_flick     & 2 & 8 & 0 & 6 & 10 & 26 \\
right\_forehand\_flick    & 0 & 1 & 0 & 0 & 0 & 1 \\
right\_backhand\_flick    & 2 & 3 & 0 & 3 & 4 & 12 \\
\hline
left\_forehand\_lob       & 1 & 2 & 1 & 0 & 0 & 4 \\
left\_backhand\_lob       & 0 & 0 & 0 & 0 & 0 & 0 \\
right\_forehand\_lob      & 0 & 2 & 0 & 1 & 0 & 3 \\
right\_backhand\_lob      & 0 & 0 & 0 & 1 & 0 & 1 \\
\hline
left\_forehand\_loop      & 17 & 58 & 24 & 17 & 12 & 128 \\
left\_backhand\_loop      & 7 & 46 & 10 & 2 & 23 & 88 \\
right\_forehand\_loop     & 24 & 31 & 24 & 16 & 14 & 109 \\
right\_backhand\_loop     & 15 & 44 & 7 & 7 & 33 & 106 \\
\hline
left\_forehand\_push      & 2 & 11 & 7 & 3 & 9 & 32 \\
left\_backhand\_push      & 4 & 16 & 4 & 24 & 38 & 86 \\
right\_forehand\_push     & 0 & 19 & 6 & 3 & 11 & 39 \\
right\_backhand\_push     & 7 & 18 & 6 & 23 & 36 & 90 \\
\hline
left\_forehand\_serve     & 8 & 25 & 13 & 6 & 18 & 70 \\
left\_backhand\_serve     & 3 & 19 & 8 & 7 & 10 & 47 \\
right\_forehand\_serve    & 6 & 43 & 2 & 11 & 22 & 84 \\
right\_backhand\_serve    & 7 & 5 & 7 & 6 & 2 & 27 \\
\hline
left\_forehand\_smash     & 1 & 2 & 0 & 2 & 0 & 5 \\
left\_backhand\_smash     & 0 & 0 & 0 & 0 & 0 & 0 \\
right\_forehand\_smash    & 1 & 2 & 0 & 0 & 0 & 3 \\
right\_backhand\_smash    & 1 & 0 & 0 & 0 & 0 & 1 \\
\bottomrule
\end{tabular}
\label{tab:apendix_train}
\end{table}

\begin{table}[h]
\centering
\caption{Strokes in the test videos}
\begin{tabular}{lcccccccc}
\toprule
\textbf{Stroke} & \textbf{Test 1} & \textbf{Test 2} & \textbf{Test 3} & \textbf{Test 4} & \textbf{Test 5} & \textbf{Test 6} & \textbf{Test 7} & \textbf{Total} \\ 
\midrule
left\_forehand\_block     & 0 & 0 & 0 & 0 & 0 & 3 & 1 & 4 \\
left\_backhand\_block     & 2 & 0 & 0 & 0 & 0 & 3 & 8 & 13 \\
right\_forehand\_block    & 0 & 0 & 0 & 0 & 0 & 0 & 0 & 0 \\
right\_backhand\_block    & 28 & 0 & 0 & 0 & 0 & 1 & 0 & 29 \\
\hline
left\_forehand\_chop      & 0 & 0 & 0 & 0 & 0 & 0 & 0 & 0 \\
left\_backhand\_chop      & 0 & 0 & 0 & 1 & 0 & 0 & 0 & 1 \\
right\_forehand\_chop     & 0 & 0 & 0 & 0 & 0 & 0 & 0 & 0 \\
right\_backhand\_chop     & 0 & 0 & 0 & 1 & 0 & 0 & 0 & 1 \\
\hline
left\_forehand\_flick     & 0 & 0 & 0 & 0 & 0 & 0 & 0 & 0 \\
left\_backhand\_flick     & 1 & 0 & 0 & 0 & 0 & 2 & 4 & 7 \\
right\_forehand\_flick    & 0 & 0 & 0 & 0 & 0 & 0 & 0 & 0 \\
right\_backhand\_flick    & 0 & 0 & 0 & 5 & 3 & 1 & 8 & 17 \\
\hline
left\_forehand\_lob       & 1 & 1 & 0 & 0 & 0 & 0 & 0 & 2 \\
left\_backhand\_lob       & 0 & 0 & 0 & 0 & 0 & 0 & 0 & 0 \\
right\_forehand\_lob      & 0 & 0 & 0 & 0 & 0 & 0 & 0 & 0 \\
right\_backhand\_lob      & 0 & 0 & 0 & 0 & 0 & 0 & 0 & 0 \\
\hline
left\_forehand\_loop      & 6 & 9 & 6 & 8 & 1 & 4 & 5 & 39 \\
left\_backhand\_loop      & 31 & 0 & 3 & 7 & 4 & 2 & 0 & 47 \\
right\_forehand\_loop     & 1 & 17 & 2 & 6 & 3 & 6 & 3 & 38 \\
right\_backhand\_loop     & 3 & 0 & 3 & 6 & 4 & 5 & 7 & 28 \\
\hline
left\_forehand\_push      & 1 & 0 & 1 & 2 & 0 & 1 & 3 & 8 \\
left\_backhand\_push      & 0 & 0 & 0 & 3 & 3 & 2 & 0 & 8 \\
right\_forehand\_push     & 1 & 0 & 2 & 1 & 1 & 0 & 1 & 6 \\
right\_backhand\_push     & 0 & 0 & 1 & 6 & 1 & 1 & 1 & 10 \\
\hline
left\_forehand\_serve     & 2 & 0 & 3 & 3 & 3 & 1 & 3 & 15 \\
left\_backhand\_serve     & 0 & 0 & 0 & 8 & 0 & 3 & 1 & 12 \\
right\_forehand\_serve    & 3 & 2 & 2 & 12 & 4 & 4 & 2 & 29 \\
right\_backhand\_serve    & 4 & 0 & 0 & 0 & 0 & 0 & 2 & 6 \\
\hline
left\_forehand\_smash     & 0 & 0 & 1 & 1 & 0 & 0 & 0 & 2 \\
left\_backhand\_smash     & 0 & 0 & 0 & 0 & 0 & 0 & 0 & 0 \\
right\_forehand\_smash    & 0 & 0 & 0 & 1 & 0 & 0 & 0 & 1 \\
right\_backhand\_smash    & 0 & 0 & 0 & 0 & 0 & 0 & 0 & 0 \\
\bottomrule
\end{tabular}
\label{tab:apendix_test}
\end{table}

\end{document}